\documentclass{article}
\usepackage{spconf,amsmath,amssymb,booktabs,multirow,graphicx,hyperref}
\title{QUERY-INDEPENDENT VARIABLE-RATE VISUAL TOKEN CODING}
\name{\fontsize{9}{10}\selectfont\begin{tabular}{@{}c@{}}
Hongbo Zhang$^{1,*}$ \quad Zihao Yang$^{2,*}$ \quad Liuyang Song$^{1}$ \quad Daqian Yang$^{1}$ \quad
Haoyang Yao$^{1}$ \quad Yan Wen$^{3}$ \quad Zhengtao Yao$^{3,\dagger}$
\end{tabular}}
\address{\fontsize{9}{10}\selectfont $^{1}$Peking University \quad $^{2}$Uppsala University \quad $^{3}$University of Southern California}
\begin{document}
\ninept
\raggedbottom
\maketitle
\begingroup
\let\thefootnote\relax
\makeatletter\long\def\@makefntext#1{\noindent#1}\makeatother
\footnotetext{\raggedright
$^{*}$Equal contribution.\\
$^{\dagger}$Corresponding author: \texttt{zyao9248@alumni.usc.edu}\par}
\endgroup

\begin{abstract}
Visual-token compression for vision--language models is posed almost entirely as
a selection problem: decide which tokens to keep and discard the rest. The
criteria that work best rank tokens by the attention the language model pays
them, which makes the ranking a function of the question being asked. That is
invisible in a single-turn benchmark and decisive whenever a compressed
representation is written once and read many times, as when it is cached across
the turns of a conversation or transmitted between a device and a server. We take the other half of the
classical transform-coding toolkit instead: keep every token and vary its rate.
A transform code exposes each token's measured distortion--rate curve, and a
fixed bit budget is distributed across tokens by exact integer rate--distortion
optimisation on those curves. No text enters the pipeline, so one compressed
representation serves any query. At equal bit budgets, on two datasets and two
capacities, it preserves the model's output distribution and its answers better
than uniform-rate coding, the closed-form water-fill and distortion-ranked
pruning. It matches attention-ranked pruning on the question
pruning was tuned for, and overtakes it once the compressed image must answer a
different question about the same image.
\end{abstract}


\section{Introduction}
Vision--language models turn an image into hundreds of visual tokens, and those
tokens are expensive to keep. Nearly all work on shrinking them asks which
tokens to discard: prune the patches the language model attends to least
\cite{fastv,prumerge,visionzip}, or merge patches that look alike \cite{tome}.
In either case the decision is binary: a token survives at full precision or it
disappears. The criteria that work best are computed from the text, because the
premise of attention-based pruning is that relevance depends on what is being
asked.
Recent work adds explicit budgets without leaving that frame: rate--utility
optimisation still gates whole tokens \cite{ruta}, learned-codebook coders reach
extreme compression at a uniform code length \cite{htcvlm}, rate
predictors adapt the budget per image over a small discrete menu
\cite{voco}, and training-free reducers push token counts down aggressively
\cite{occam,fourier}. None varies the number of bits spent on a surviving token.

Conditioning on the query is invisible in the usual single-turn benchmark and
decisive outside it. A visual representation is often written once and read
repeatedly: cached across the turns of a conversation, split between a device
that runs the vision tower and a server that runs the language model, or stored
for a corpus queried many ways. The encoder then does not know the question, and
a ranking computed from one question is the wrong ranking for the next.

We therefore take the half of the classical transform-coding toolkit this
literature has left unused: keep every token and vary its \emph{rate}. Under a
fixed bit budget the rate--distortion-optimal allocation across a set of sources
is generally not binary. Reverse water-filling gives each source a rate that
grows with its variance \cite{cover}, and keep/drop is the degenerate two-valued
case. When each source instead exposes a finite set of quantisers with
\emph{measured} rate--distortion pairs, the allocation becomes a separable
resource problem that dynamic programming solves exactly \cite{shoham}. We keep
that machinery and change what it is applied to, the visual-token embeddings a
frozen model consumes rather than pixels destined for a human viewer, and we
take the distortion--rate curves from measurement rather than from a parametric
model. Because those curves are a property of the image, the resulting
allocation is query-independent by construction.

We contribute three things. The first is a query-independent variable-rate
visual-token codec whose allocation is solved exactly on measured
distortion--rate curves, with no tuned parameter. The second is a set of gains
over uniform-rate coding and over pruning at equal bits, consistent across two
datasets and two model capacities. The third is a transfer protocol. Every
ChartQA test image carries two independent human questions, which lets us
measure what conditioning an allocation on the query costs once the
representation outlives that query.

\begin{figure*}[t]
\centering
\includegraphics[width=\textwidth]{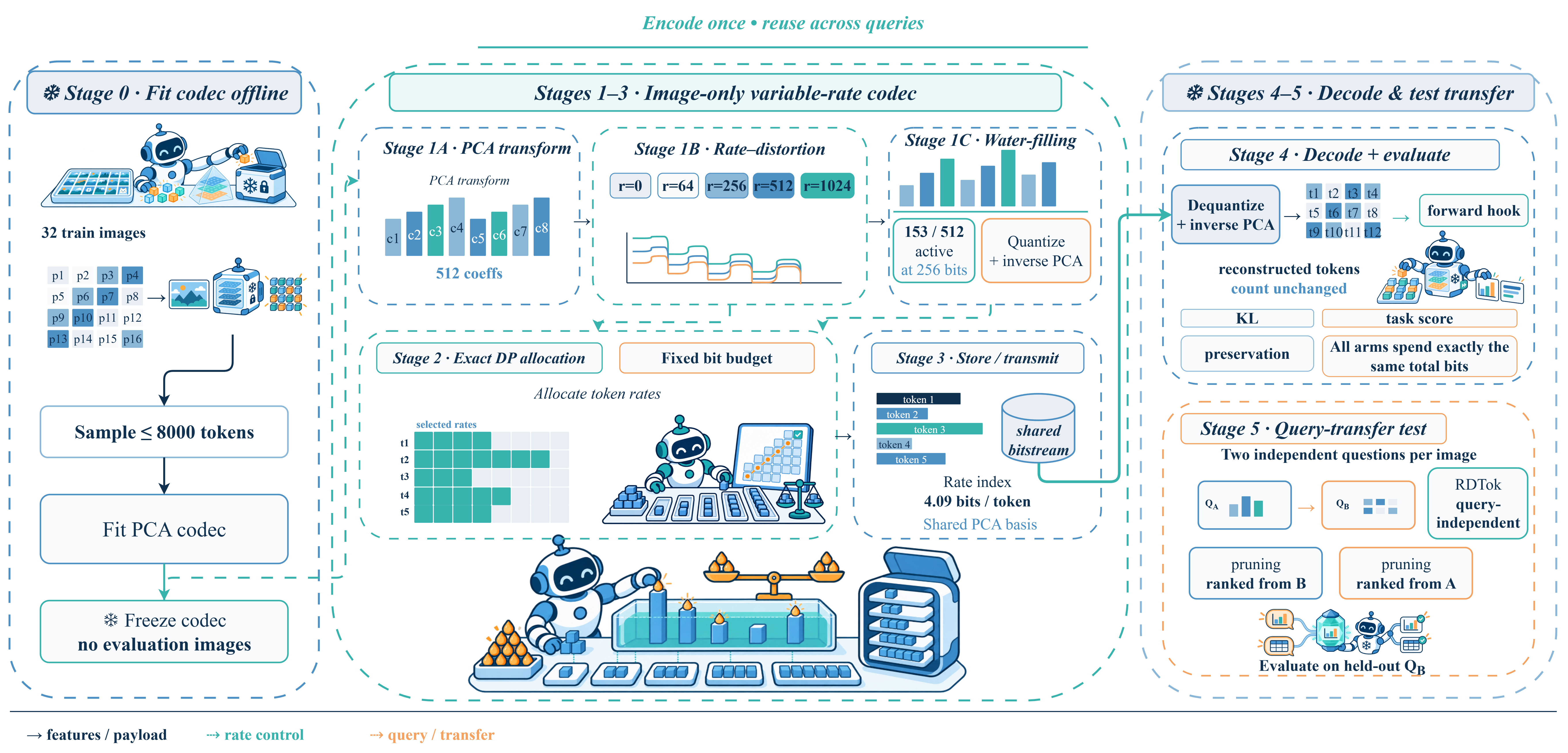}
\caption{RDTok. The codec measures each token's distortion--rate curve and
distributes a fixed bit budget across tokens by exact dynamic programming. No
query enters the encoder, so one coded representation is written once and reused
by any later query. The token count is unchanged; what varies is the bits each
token receives.}
\label{fig:framework}
\end{figure*}

\section{Method}
\subsection{Transform and quantiser}
We call the resulting codec RDTok. It has two parts: a transform code that turns
each token into a measured distortion--rate curve, and an allocator that spends
a fixed bit budget across those curves. Fig.~\ref{fig:framework} gives the
pipeline. Let $z_i\in\mathbb R^{d}$, $i=1,\dots,N$, be the visual tokens emitted by the
frozen vision tower ($d=2560$ for Qwen3-VL). On tokens from images disjoint from
every evaluation image we fit a mean $\mu$ and an orthonormal basis
$U\in\mathbb R^{K\times d}$, $K=512$, and code each token in the coefficient
domain, $y_i=U(z_i-\mu)$, $\hat z_i=U^{\!\top}\hat y_i+\mu$. By orthonormality
the PCA truncation error is independent of the rate, so it is identical for
every allocator and cancels in all comparisons below.

Let $\sigma_j^2$ be the variance of coefficient $j$ on the fitting split. Under
the high-resolution model $D_j(b)\propto\sigma_j^{2}2^{-2b}$, spending $r$ bits
on a token is minimised by the classical reverse water-fill \cite{cover},
\begin{equation}
b_j(r)=\Big[\tfrac12\log_2\!\big(\sigma_j^{2}/\theta(r)\big)\Big]_{+},
\qquad \sum\nolimits_{j} b_j(r)=r ,
\label{eq:waterfill}
\end{equation}
with the water level $\theta(r)$ obtained by bisection; the clipped sum is
continuous and monotone in $-\log\theta$, so the budget is met exactly. Each
coefficient then receives $M_j=\max(\lfloor 2^{b_j}\rceil,1)$ levels and a
uniform mid-tread quantiser over $\pm c\sigma_j$, $c=4$:
\begin{equation}
\Delta_j=\frac{2c\,\sigma_j}{M_j},\qquad
\hat y_{ij}=\Delta_j\,\mathrm{clip}\!\left(
\Big\lfloor \frac{y_{ij}}{\Delta_j}\Big\rceil,\
-\tfrac{M_j-1}{2},\ \tfrac{M_j-1}{2}\right),
\end{equation}
where $M_j=1$ decodes to the PCA mean. Eq.~\eqref{eq:waterfill} therefore
also selects the active subspace: at $256$ bits only $153$ of the $512$
coefficients are funded, rising to all $512$ at $1024$.

\begin{figure}[tb]
\centering
\includegraphics[width=\columnwidth]{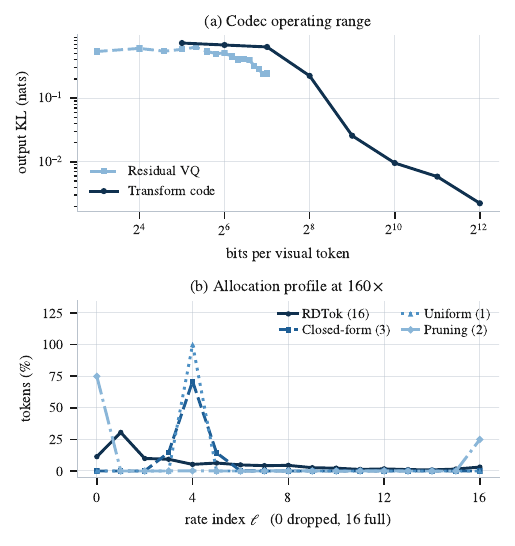}
\caption{(a) Output KL against per-token rate for the two codec families on
ChartQA. (b) Tokens placed at each rate index, at $160\times$; the legend gives
the number of levels carrying $99\%$ of the tokens.}
\label{fig:main}
\end{figure}

Rates lie on a uniform grid $r=\ell\rho$, $\ell\in\mathcal L=\{0,\dots,L\}$,
with $L=16$ and $\rho=64$ bits, so a rate index is interchangeable with a bit
count and $\ell=0$ is the dropped-token analogue. One pass over the grid yields
\begin{equation}
D_i(\ell)=\big\lVert z_i-\hat z_i(\ell\rho)\big\rVert^{2},
\qquad \ell\in\mathcal L,
\end{equation}
the \emph{measured} distortion--rate curve of token $i$. These curves, not a
parametric model of them, are what the allocator consumes. A transform code is
used rather than a learned vector quantiser because it has an operating range.
A shared-codebook residual VQ saturates near $0.25$ nats and is non-monotone in
rate, whereas the transform code spans four orders of magnitude and is nearly
lossless at full rate (Fig.~\ref{fig:main}a). The budget, not the codec, then
creates the distortion being allocated.

\subsection{Allocation across tokens}
An allocator is a map $\mathcal A:(D,s,B)\mapsto\ell\in\mathcal L^{N}$
with $\sum_i\ell_i=B$, where $D$ are the measured curves and $s$ an optional
per-token score. Every method compared here is such a map with the same $B$,
which makes the comparison equal-rate by construction. Ours uses no $s$:
\begin{equation}
\ell^{\star}=\arg\min_{\ell\in\mathcal L^{N}} \sum_{i=1}^{N} D_i(\ell_i)
\quad\text{s.t.}\quad \sum\nolimits_{i}\ell_i=B=\beta NL .
\label{eq:alloc}
\end{equation}
The objective is separable and the constraint is one scalar resource, so
Eq.~\eqref{eq:alloc} is solved exactly by dynamic programming over the spent
budget \cite{shoham}. With $J_k(b)$ the least distortion achievable by the first
$k$ tokens spending exactly $b$ steps,
\begin{equation}
J_k(b)=\min_{\ell\in\mathcal L,\ \ell\le b}\Big\{J_{k-1}(b-\ell)+D_k(\ell)\Big\},
\label{eq:dp}
\end{equation}
with $J_0(0)=0$ and $J_0(b)=\infty$ otherwise; the optimum is $J_N(B)$ and
backtracking the minimising $\ell$ recovers the allocation, ties broken toward
the smaller $\ell$ for determinism. Unreachable states are skipped, leaving
$O(NBL)$ time; one solve takes $58$\,ms on average, encoder-side only.

Solving exactly, rather than substituting the closed form of
Eq.~\eqref{eq:waterfill} across tokens, matters for two reasons. The measured
$D_i$ obey neither the parametric decay that form assumes nor convexity: only
$59.6\%$ of marginal-gain differences are non-increasing. The closed form is
also not scale invariant, so it needs a normalised importance, a knob
Eq.~\eqref{eq:dp} does not have.

\noindent\textbf{Baselines as allocators.} Uniform rate sets
$\ell_i=\lfloor B/N\rceil$ and the closed form substitutes
$s_i=D_i(0)-D_i(L)$ into Eq.~\eqref{eq:waterfill}, both rounded onto
$\mathcal L^{N}$ with exact sum $B$. Pruning ranks tokens by a score $s$ and,
with $k=\lfloor B/L\rfloor$ and $\pi$ the induced ordering,
\begin{equation}
\ell_{\pi(m)}=L \ (m\le k),\quad B-kL \ (m=k{+}1),\quad 0 \ \text{otherwise},
\label{eq:prune}
\end{equation}
which is exactly the two-valued restriction $\ell_i\in\{0,L\}$ of
Eq.~\eqref{eq:alloc}---a feasible but generally suboptimal point of the same
programme. It is instantiated with the source score $s_i=D_i(0)-D_i(L)$ and with
the attention score $s_i=\sum_{t\in\mathcal T}\bar a_{t\to i}$, the mass text
positions $\mathcal T$ send to token $i$ at a mid-depth layer, which is the
token-pruning criterion \cite{fastv,prumerge,visionzip}.

\noindent\textbf{Rate accounting and query independence.} Each token stores its
rate index, $\log_2(L+1)=4.09$ bits, $1.6\%$ of the budget at $256$ bits per
token; scalar quantisation needs no codebook and $U$ amortises over the corpus.
Tables report the coefficient budget, identical across arms. No text enters
Eqs.~\eqref{eq:waterfill}--\eqref{eq:dp}: the vision tower does not see the
prompt, so $D$ and $\ell^{\star}$ are functions of the image alone. Among the
baselines only pruning with the attention score reads the query.

\subsection{Why the objective stays on the source side}
The evident refinement is to weight Eq.~\eqref{eq:alloc} by each region's
measured effect on the output: partition the tokens into $G$ contiguous bands,
spend one forward pass per band measuring the output KL $\kappa_g$ caused by
dropping band $\mathcal B_g$, and minimise $\sum_i w_{g(i)}D_i(\ell_i)$ with
$w_g=\kappa_g/\sum_{i\in\mathcal B_g}D_i(0)$, which is Eq.~\eqref{eq:dp} on
weighted curves. Such a weighting measures something real---feature MSE is not a
sufficient statistic for output KL (Sec.~\ref{sec:ablation})---but what it
measures is a property of the \emph{query} it came from, not of the image, and
it does not transfer to the next question about the same image.
Sec.~\ref{sec:transfer} shows the attention score of Eq.~\eqref{eq:prune} has
the same defect, which makes this a fact about query-conditioned allocation as a
class rather than about one signal.

\begin{figure*}[t]
\centering
\includegraphics[width=\textwidth]{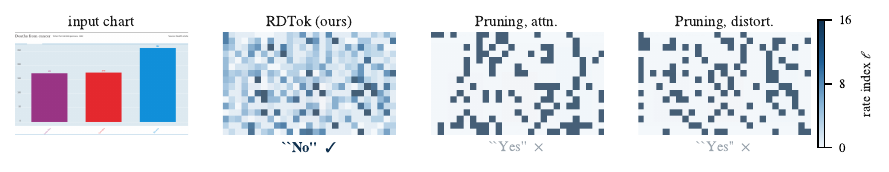}
\caption{A ChartQA case at $160\times$ on Qwen3-VL-8B (reference \emph{No}, which
the uncompressed model gives). Panels show the rate index each allocator assigns
to each token, on the token grid. Only RDTok uses intermediate levels, and only
RDTok still answers correctly.}
\label{fig:qualitative}
\end{figure*}

\begin{figure}[tb]
\centering
\includegraphics[width=\columnwidth]{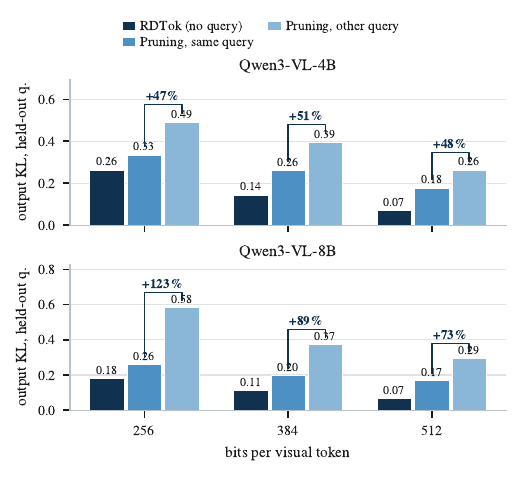}
\caption{Query transfer on ChartQA ($n=288$ images with two human questions
each; all arms scored on the held-out question, mean output KL). Attention-ranked
pruning appears twice: ranked on the question it is scored on, and on the
\emph{other} question of that image. The brace is the degradation.}
\label{fig:transfer}
\end{figure}

\section{Experiments}
Three questions organise the evaluation: whether varying the rate helps at all
at a fixed bit budget, whether it helps more than selecting tokens, and what
happens when the compressed representation has to serve a question the encoder
never saw.

\subsection{Protocol}
Backbones are the 4B and 8B Qwen3-VL-Instruct models \cite{qwenvl}, frozen, in
float32; datasets are ChartQA \cite{chartqa} (relaxed accuracy) and COCO
captions \cite{coco}, $n=288$ evaluation images per cell, with the basis and
quantiser fitted on 32 images from the dataset's own train split. The three rates,
$256/384/512$ bits per token, are $160\times$, $106\times$ and $80\times$
compression of the fp16 features; they were chosen per backbone, by the same
rule for both, as the band where a measured rate--KL curve is monotone and
unsaturated. All arms spend the same total bits. Significance is exact McNemar for the binary
metrics and Wilcoxon signed-rank for KL; $^{\ast}$ marks $p<0.05$ against RDTok.

\begin{table}[tb]
\caption{ChartQA, $n=288$ per cell, equal bit budget; COCO is in the text.
Uncompressed accuracy $75.3$ (4B) / $76.7$ (8B). ``Pres.'': output preservation.
Best per column in bold; $^{\ast}$: $p<0.05$ against RDTok.}
\label{tab:main}
\centering
\setlength{\tabcolsep}{2.2pt}
\resizebox{\columnwidth}{!}{%
\begin{tabular}{lrrrrrr}
\toprule
& \multicolumn{3}{c}{Qwen3-VL-4B} & \multicolumn{3}{c}{Qwen3-VL-8B} \\
\cmidrule(lr){2-4}\cmidrule(lr){5-7}
Method & KL$\downarrow$ & Pres.$\uparrow$ & Acc.$\uparrow$ & KL$\downarrow$ & Pres.$\uparrow$ & Acc.$\uparrow$ \\
\midrule
\multicolumn{7}{l}{\emph{256 bits/token ($160\times$)}} \\
\quad \textbf{RDTok (ours)} & \textbf{0.219} & \textbf{68.4} & 63.2 & \textbf{0.166} & \textbf{72.9} & 70.5 \\
\quad Closed-form water-fill & 0.287$^{\ast}$ & 56.2$^{\ast}$ & 54.2$^{\ast}$ & 0.285$^{\ast}$ & 61.1$^{\ast}$ & 63.2$^{\ast}$ \\
\quad Uniform rate & 0.311$^{\ast}$ & 54.9$^{\ast}$ & 52.4$^{\ast}$ & 0.327$^{\ast}$ & 61.5$^{\ast}$ & 60.1$^{\ast}$ \\
\quad Pruning, distortion-rank & 0.510$^{\ast}$ & 46.9$^{\ast}$ & 50.0$^{\ast}$ & 0.432$^{\ast}$ & 50.3$^{\ast}$ & 52.4$^{\ast}$ \\
\quad Pruning, attention-rank & 0.324$^{\ast}$ & 61.1$^{\ast}$ & \textbf{63.5} & 0.262 & 68.1 & \textbf{70.8} \\
\addlinespace
\multicolumn{7}{l}{\emph{384 bits/token ($106\times$)}} \\
\quad \textbf{RDTok (ours)} & \textbf{0.106} & \textbf{76.0} & \textbf{68.1} & \textbf{0.077} & \textbf{80.6} & \textbf{75.3} \\
\quad Closed-form water-fill & 0.143$^{\ast}$ & 67.4$^{\ast}$ & 62.5$^{\ast}$ & 0.149$^{\ast}$ & 75.7$^{\ast}$ & 71.5$^{\ast}$ \\
\quad Uniform rate & 0.160$^{\ast}$ & 68.1$^{\ast}$ & 60.8$^{\ast}$ & 0.148$^{\ast}$ & 72.9$^{\ast}$ & 70.5$^{\ast}$ \\
\quad Pruning, distortion-rank & 0.392$^{\ast}$ & 59.4$^{\ast}$ & 58.0$^{\ast}$ & 0.239$^{\ast}$ & 70.5$^{\ast}$ & 67.7$^{\ast}$ \\
\quad Pruning, attention-rank & 0.216$^{\ast}$ & 70.1$^{\ast}$ & 68.1 & 0.196$^{\ast}$ & 74.3$^{\ast}$ & 71.2 \\
\addlinespace
\multicolumn{7}{l}{\emph{512 bits/token ($80\times$)}} \\
\quad \textbf{RDTok (ours)} & \textbf{0.064} & \textbf{80.2} & \textbf{69.8} & \textbf{0.059} & \textbf{85.1} & \textbf{76.4} \\
\quad Closed-form water-fill & 0.081$^{\ast}$ & 75.7$^{\ast}$ & 67.7 & 0.072$^{\ast}$ & 81.9 & 74.3 \\
\quad Uniform rate & 0.084$^{\ast}$ & 75.3$^{\ast}$ & 68.8 & 0.082$^{\ast}$ & 81.6 & 76.0 \\
\quad Pruning, distortion-rank & 0.162$^{\ast}$ & 70.5$^{\ast}$ & 66.3 & 0.076$^{\ast}$ & 80.2$^{\ast}$ & 75.7 \\
\quad Pruning, attention-rank & 0.167$^{\ast}$ & 74.7$^{\ast}$ & 69.4 & 0.166$^{\ast}$ & 78.1$^{\ast}$ & 73.6 \\
\addlinespace
\bottomrule
\end{tabular}%
}
\end{table}

\subsection{Main results}
Table~\ref{tab:main} reports ChartQA at both capacities. Against every query-independent baseline, RDTok lowers output KL in every cell
and leads on the other two quantities in all but a couple of COCO cells, and the
margins widen as compression becomes more aggressive.
That ordering follows from what each arm can express. Uniform rate ignores the
heterogeneity of the per-token curves entirely, being the single-level point of
Eq.~\eqref{eq:alloc}, so it overspends on tokens whose distortion has already
flattened and underspends on those still descending steeply. Distortion-ranked
pruning is the opposite failure. Being the two-valued restriction
$\ell_i\in\{0,L\}$ of the same programme, it commits its whole budget to a
quarter of the tokens and discards the rest outright, when a few levels each
would have retained most of their content at a fraction of the cost. Both are
feasible points of Eq.~\eqref{eq:alloc}, which is why solving it properly
dominates them rather than trading against them.

The closed-form water-fill is the informative negative control, because it also
varies the rate and still lands near uniform rate. Fig.~\ref{fig:main}b explains
why: assuming a parametric decay the measured curves do not follow collapses its
allocation onto three levels, close to the single level uniform rate uses. The
gain therefore comes from taking the distortion--rate curves from
\emph{measurement}, not from varying the rate as such. That is the one design decision
the paper most depends on.

The second dataset behaves the same way at both capacities. Across the three
rates on COCO captions, RDTok lowers output KL against uniform rate by
$37$--$47\%$ at 4B and $43$--$45\%$ at 8B, against the closed-form water-fill by
$34$--$43\%$ and $40$--$43\%$, and against distortion-ranked pruning by
$15$--$27\%$ and $9$--$37\%$. Every comparison reaches $p<10^{-3}$, and on 8B
RDTok also beats attention-ranked pruning at the most aggressive rate.

Over three seeds at each capacity, each redrawing both splits, RDTok reaches
$64.2\pm0.9$ and $68.6\pm2.6$ against $55.6\pm3.0$ and $57.9\pm1.9$ for
uniform rate.
The ordering is unchanged across those seeds, and across two datasets as
different as dense synthetic charts and natural photographs. It does not weaken
at the larger capacity either. Against pruning it strengthens, the opposite of
the usual pattern for a method developed on a small model.

Attention-ranked pruning is the strongest baseline and the informative
comparison. Reading the query, it selects tokens about as well as rate
allocation spends bits, matching RDTok on task accuracy, while RDTok, reading
only the image, leads it on output KL and on output preservation. Parity with a
query-conditioned method at no query cost is already the useful result; it
becomes decisive once the compressed representation has to outlive the query.
Fig.~\ref{fig:qualitative} shows one such case: the pruning arms discard three
quarters of the tokens and answer wrongly, while RDTok spreads the same budget
over sixteen levels and reproduces the uncompressed answer.

\subsection{Query transfer}
\label{sec:transfer}
Every ChartQA test image comes with two independent human questions, so a
pruning ranking can be computed from question A and scored on question B, the
situation of a cached or transmitted image. Fig.~\ref{fig:transfer} reports it.

Carried to the other question, attention-ranked pruning loses roughly half its
fidelity at the smaller capacity and more than doubles its divergence at the
larger one, and RDTok leads it at every rate and both capacities. The asymmetry
is structural rather than empirical. An attention score marks the regions the
\emph{current} question needs; a distortion--rate curve marks what the
\emph{image} contains. The first is a property of the query and has no reason to
survive a change of query; the second is a property of the image and is
unchanged by one. A query-independent allocator therefore has no transfer term
to lose, which is why its advantage appears here and not in the single-turn
comparison above.

The larger model's attention is the more query-specific, so a larger share of what
pruning selects is particular to the question asked, and the value of not
conditioning rises with the model. Query-aware selection is thus a trade of
reusability for per-query fit, and reusability is precisely what the
write-once-read-many setting is made of.

\subsection{Analysis}
\label{sec:ablation}
\noindent\textbf{What the allocation looks like.} Fig.~\ref{fig:main}b makes the
difference from pruning concrete. RDTok spreads the budget over sixteen levels,
pruning has two by construction and uniform rate one. The closed-form water-fill
collapses onto three, which is why it tracks uniform rate so closely in
Table~\ref{tab:main} despite also varying the rate, and why solving on measured
curves rather than a parametric model is where the gain comes from.

\noindent\textbf{Why the source objective is not the whole story.} Perturbations
carrying identical total squared error but different shapes separate the output
KL by $2.9$ to $14\times$ at a fixed error level; if feature MSE determined
output KL they would coincide. Concentrating the same error on a quarter of the
tokens is several times worse than spreading it, and the worst shape is always
the high-norm tokens or the leading PCA directions. A source-MSE allocator is therefore optimising a surrogate, which bounds what
any allocator working from $D$ alone can achieve, and is why the query-weighted
variant of Sec.~2.3 was worth trying at all.

\end{document}